\documentclass{article}

\usepackage{microtype}
\usepackage{graphicx}
\usepackage{subfigure}
\usepackage{amsmath}
\usepackage{multirow}
\usepackage{booktabs} %

\usepackage[breaklinks=true]{hyperref}

\usepackage[accepted]{sty/icml2019}

\begin{document}

\twocolumn[
\icmltitle{EfficientNet: Rethinking Model Scaling for Convolutional Neural Networks}
\icmlsetsymbol{equal}{*}

\begin{icmlauthorlist}
\icmlauthor{Mingxing Tan}{brain}
\icmlauthor{Quoc V. Le}{brain}
\end{icmlauthorlist}

\icmlaffiliation{brain}{Google Research, Brain Team, Mountain View, CA}

\icmlcorrespondingauthor{Mingxing Tan}{tanmingxing@google.com}

\icmlkeywords{Machine Learning, ICML}

\vskip 0.3in
]

\printAffiliationsAndNotice{}  %

\def\TODO{\textcolor{red}{\emph{TODO: }}}
\newcommand{\TT}[1]{\texttt{#1}}
\newcommand{\BF}[1]{\textbf{#1}}
\newcommand{\IT}[1]{\textit{#1}}

\newcommand\blfootnote[1]{%
	\begingroup
	\renewcommand\thefootnote{}\footnote{#1}%
	\addtocounter{footnote}{-1}%
	\endgroup
}

\newcommand{\km}[1]{{\color{red}[km: #1]}}                                          
\newcommand{\rbg}[1]{{\color{blue}[rbg: #1]}}                                       
\newcommand{\ppd}[1]{{\color{green}[ppd: #1]}}                                      
\newcommand{\bd}[1]{\textbf{#1}}                                                    
\newcommand{\app}{\raise.17ex\hbox{$\scriptstyle\sim$}}                             
\newcommand{\symb}[1]{{\small\texttt{#1}}\xspace}                                   
\newcommand{\mrtwo}[1]{\multirow{2}{*}{#1}}                                         
\def\x{\times}                                                                      
\def\pt{p_\textrm{t}}                                                               
\def\at{\alpha_\textrm{t}}                                                          
\def\xt{x_\textrm{t}}                                                               
\def\CE{\textrm{CE}}                                                                
\def\FL{\textrm{FL}}                                                                
\def\FQ{\textrm{FL}^*}                                                              
\newcommand{\eqnnm}[2]{\begin{equation}\label{eq:#1}#2\end{equation}\ignorespaces}

\newlength\savewidth\newcommand\shline{\noalign{\global\savewidth\arrayrulewidth 
		\global\arrayrulewidth 1pt}\hline\noalign{\global\arrayrulewidth\savewidth}}   
\newcommand{\tablestyle}[2]{\setlength{\tabcolsep}{#1}\renewcommand{\arraystretch}{#2}\centering\footnotesize}
\makeatletter\renewcommand\paragraph{\@startsection{paragraph}{4}{\z@}              
	{.5em \@plus1ex \@minus.2ex}{-.5em}{\normalfont\normalsize\bfseries}}\makeatother
\renewcommand{\dbltopfraction}{1}                                                   
\renewcommand{\bottomfraction}{0}                                                   
\renewcommand{\textfraction}{0}                                                     
\renewcommand{\dblfloatpagefraction}{0.95}                                          
\setcounter{dbltopnumber}{5}

\newcommand{\M}[1]{\mathcal{#1}}
\begin{abstract}

Convolutional Neural Networks (ConvNets) are commonly developed at a fixed resource budget, and then scaled up for better accuracy if more resources are available.  In this paper, we systematically study model scaling and identify that carefully balancing network depth, width, and resolution can lead to better performance. Based on this observation, we propose a new scaling method that uniformly scales all dimensions of depth/width/resolution using a simple yet highly effective \emph{compound coefficient}. We demonstrate the effectiveness of this method on scaling up MobileNets and ResNet.

To go even further, we use neural architecture search to design a new baseline network and scale it up to obtain a family of models, called \emph{EfficientNets}, which achieve much better accuracy and efficiency than previous ConvNets. In particular, our EfficientNet-B7 achieves state-of-the-art 84.3\% top-1 accuracy on ImageNet, while being \BF{8.4x smaller} and \BF{6.1x faster} on inference than the best existing ConvNet. Our EfficientNets also transfer well and achieve state-of-the-art accuracy on CIFAR-100 (91.7\%), Flowers (98.8\%), and 3 other transfer learning datasets, with an order of magnitude fewer parameters. Source code is at {\footnotesize \url{https://github.com/tensorflow/tpu/tree/master/models/official/efficientnet}}.

\end{abstract}
\begin{figure}[t]
	\includegraphics[width=0.99\columnwidth]{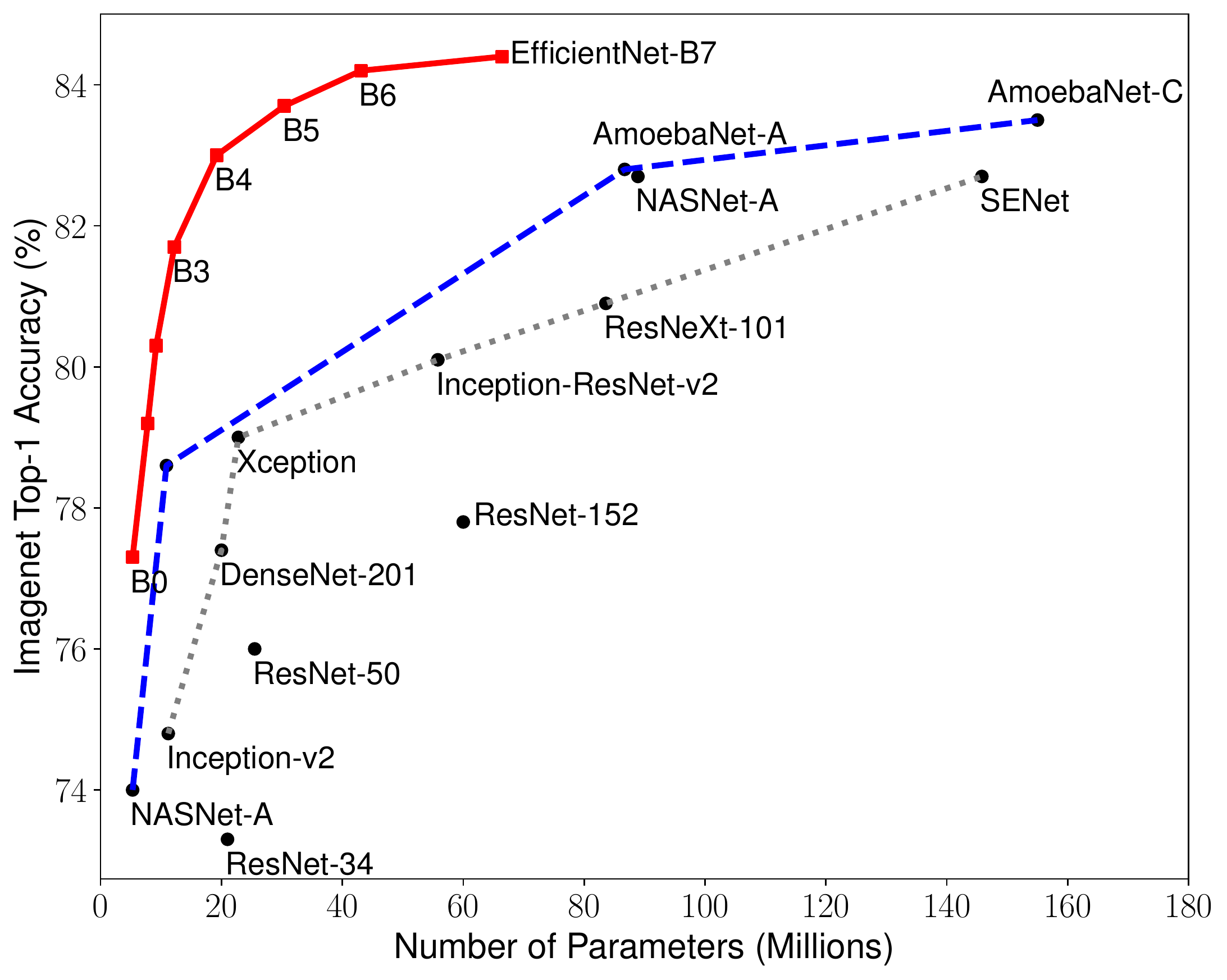}
	\hspace{-44mm}\resizebox{.46\columnwidth}{!}{\tablestyle{2pt}{1}
		\begin{tabular}[b]{l|cr}
			& Top1 Acc. & \#Params  \\
			\shline
			ResNet-152 \cite{resnet16} & 77.8\%  & 60M \\
			\bf EfficientNet-B1 & \bf 79.1\%  & \bf 7.8M \\
			\hline
			ResNeXt-101 \cite{resnext17}         &  80.9\% & 84M \\
			\bf EfficientNet-B3                                    & \bf 81.6\%  & \bf 12M \\
			\hline
			SENet  \cite{senet18}                       &  82.7\% & 146M \\
			NASNet-A \cite{nas_imagenet18}     & 82.7\% & 89M \\
            \bf EfficientNet-B4                                    & \bf 82.9\% & \bf 19M \\
			\hline
			GPipe \cite{gpipe18}  $^\dagger$  &  84.3\% & 556M \\
			\bf EfficientNet-B7                                    & \bf 84.3\% & \bf 66M \\
			\multicolumn{3}{l}{~~$^\dagger$Not plotted ~~\vspace{15mm} } \\
	\end{tabular}}
    \vskip -0.15in
	\caption{\BF{Model Size vs. ImageNet Accuracy.} All numbers are for single-crop, single-model. Our EfficientNets significantly outperform other ConvNets. In particular,  EfficientNet-B7 achieves new state-of-the-art 84.3\% top-1 accuracy but being 8.4x smaller and 6.1x faster than GPipe. EfficientNet-B1 is 7.6x smaller and 5.7x faster than ResNet-152. Details are in Table \ref{tab:imagenet} and \ref{tab:latency}.}
	\label{fig:imagnet-params}
\end{figure}
\begin{figure*}                                                          
        \centering
        \includegraphics[width=0.97\textwidth]{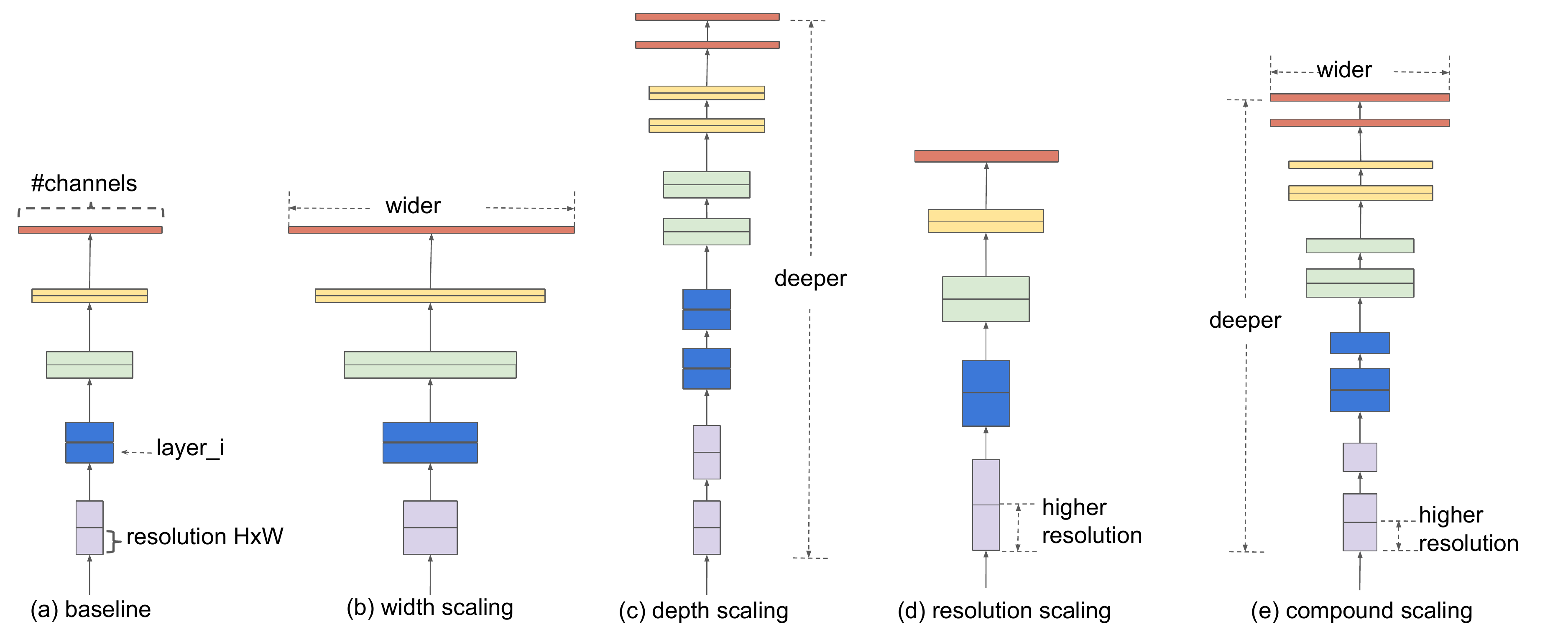} 
        \caption{\textbf{Model Scaling.} (a) is a baseline network example; (b)-(d) are conventional scaling that only increases one dimension of  network width, depth, or resolution. (e) is our proposed compound scaling method that uniformly scales all three dimensions with a fixed ratio.}                                                                        
        \label{fig:scalecompare}   

\end{figure*}  
 \section{Introduction}
Scaling up ConvNets is widely used to achieve better accuracy. For example, ResNet~\cite{resnet16} can be scaled up from
ResNet-18 to ResNet-200 by using more layers;  Recently, GPipe~\cite{gpipe18} achieved 84.3\% ImageNet top-1 accuracy by scaling up a baseline model four time larger. 
However, the process of  scaling up ConvNets has never been well understood and there are currently many ways to do it. The most common way is to scale up ConvNets by their depth~\cite{resnet16} or width~\cite{wideresnet16}. Another less common, but increasingly popular, method is to scale up models by image resolution ~\cite{gpipe18}.  In previous work, it is common to scale only one of the three dimensions -- depth, width, and image size. Though it is possible to scale  two or three dimensions  arbitrarily, arbitrary scaling requires tedious manual tuning and still often yields sub-optimal accuracy and efficiency.

In this paper, we want to study and rethink the process of scaling up ConvNets. In particular, we investigate the central question: is there a principled method to scale up ConvNets that can achieve better accuracy and efficiency? 
Our empirical study shows that it is critical to  balance all dimensions of network width/depth/resolution, and surprisingly such balance can be achieved by simply scaling each of them with constant ratio. Based on this observation, we propose a simple yet effective \emph{compound scaling method}.  Unlike conventional practice that arbitrary scales  these factors, our method uniformly scales  network width, depth, and resolution with a set of fixed scaling coefficients. For example, if we want to use $2^N$ times more computational resources, then we can simply increase the network depth by $\alpha ^ N$,  width by $\beta ^ N$, and image size by $\gamma ^ N$, where $\alpha, \beta, \gamma$ are constant coefficients determined by a small grid search on the original small model. Figure \ref{fig:scalecompare} illustrates the difference between our  scaling method and conventional methods.

Intuitively, the compound scaling method makes sense because if the input image is bigger, then the network needs more layers to increase the receptive field and more channels to capture more fine-grained patterns on the bigger image. In fact, previous theoretical  \cite{expressdepth17,expresswidth18} and empirical results \cite{wideresnet16} both show that there exists certain relationship between network width and depth, but to our best knowledge, we are the first to empirically quantify the relationship among all three dimensions of network width, depth, and resolution.

We demonstrate that our scaling method work well on existing MobileNets \cite{mobilenetv117, mobilenetv218} and ResNet \cite{resnet16}. Notably, the effectiveness of model scaling heavily depends on the baseline network; to go even further, we  use neural architecture search \cite{nas_cifar17,mnas18} to develop a new  baseline network, and scale it up to obtain a family of models, called \emph{EfficientNets}. Figure \ref{fig:imagnet-params} summarizes the ImageNet performance, where our EfficientNets  significantly outperform other ConvNets. In particular, our EfficientNet-B7  surpasses the best existing GPipe accuracy \cite{gpipe18}, but using 8.4x fewer parameters and running 6.1x faster on inference. Compared to the widely used ResNet-50 \cite{resnet16}, our EfficientNet-B4 improves the top-1 accuracy from 76.3\% to 83.0\% (+6.7\%) with similar FLOPS. Besides ImageNet,  EfficientNets also transfer well and achieve state-of-the-art accuracy on 5 out of 8 widely used datasets, while reducing parameters by up to 21x than existing ConvNets.  %
\section{Related Work}                                                           
 \label{sec:related}
 
\paragraph{ConvNet Accuracy: } Since AlexNet \cite{alexnet12} won the 2012 ImageNet competition, ConvNets have become increasingly more accurate by going bigger: while the 2014 ImageNet winner GoogleNet \cite{googlenet14} achieves 74.8\% top-1 accuracy with about 6.8M parameters,  the 2017 ImageNet winner SENet \cite{senet18} achieves 82.7\% top-1 accuracy with 145M parameters. Recently, GPipe \cite{gpipe18} further pushes the state-of-the-art ImageNet top-1 validation accuracy to 84.3\% using 557M parameters:  it is so big that it can only be trained with a specialized pipeline parallelism library by partitioning the network and spreading each part to a different accelerator. While these models are mainly designed for ImageNet, recent studies have shown better ImageNet models also perform better across a variety of transfer learning datasets \cite{imagenettransfer18}, and other computer vision tasks such as object detection \cite{resnet16,mnas18}. Although higher accuracy is critical for many applications, we have already hit the hardware memory limit, and thus further accuracy gain needs better efficiency.

\paragraph{ConvNet Efficiency: } Deep ConvNets are often over-parameterized. Model compression \cite{quantize15,amc18,netadapt18} is a common way to reduce model size by trading accuracy for efficiency. As mobile phones become ubiquitous, it is also  common to hand-craft efficient mobile-size ConvNets, such as SqueezeNets \cite{squeezenet16,squeezeNext18}, MobileNets \cite{mobilenetv117,mobilenetv218}, and ShuffleNets \cite{shufflenet17,shufflenetv218}. Recently, neural architecture search becomes increasingly popular in designing efficient mobile-size ConvNets \cite{mnas18,proxyless18}, and achieves even better efficiency than hand-crafted mobile ConvNets by extensively tuning the network width, depth, convolution kernel types and sizes. However, it is  unclear how to apply these techniques for larger models that have much larger design space and much more expensive tuning cost. In this paper, we aim to study model efficiency for super large ConvNets that surpass state-of-the-art accuracy. To achieve this goal, we resort to model scaling.
 
\paragraph{Model Scaling: } There are many ways to scale a ConvNet for different resource constraints: ResNet \cite{resnet16} can be scaled down (e.g., ResNet-18) or up (e.g., ResNet-200) by adjusting network depth (\#layers), while WideResNet \cite{wideresnet16} and MobileNets \cite{mobilenetv117} can be scaled by network width (\#channels).  It is also well-recognized that bigger input image size will help accuracy with the overhead of more FLOPS.  Although prior studies \cite{expressdepth17,expressoneneuron18,expressoverlap18,expresswidth18} have shown that network depth and width are both important for ConvNets' expressive power, it still remains an open question of how to effectively scale a ConvNet to achieve better efficiency and accuracy. Our work systematically and empirically studies ConvNet scaling for all three dimensions of network width, depth, and resolutions.  %
\begin{figure*}                                                          
        \centering                                                                  
        \includegraphics[width=0.98\textwidth]{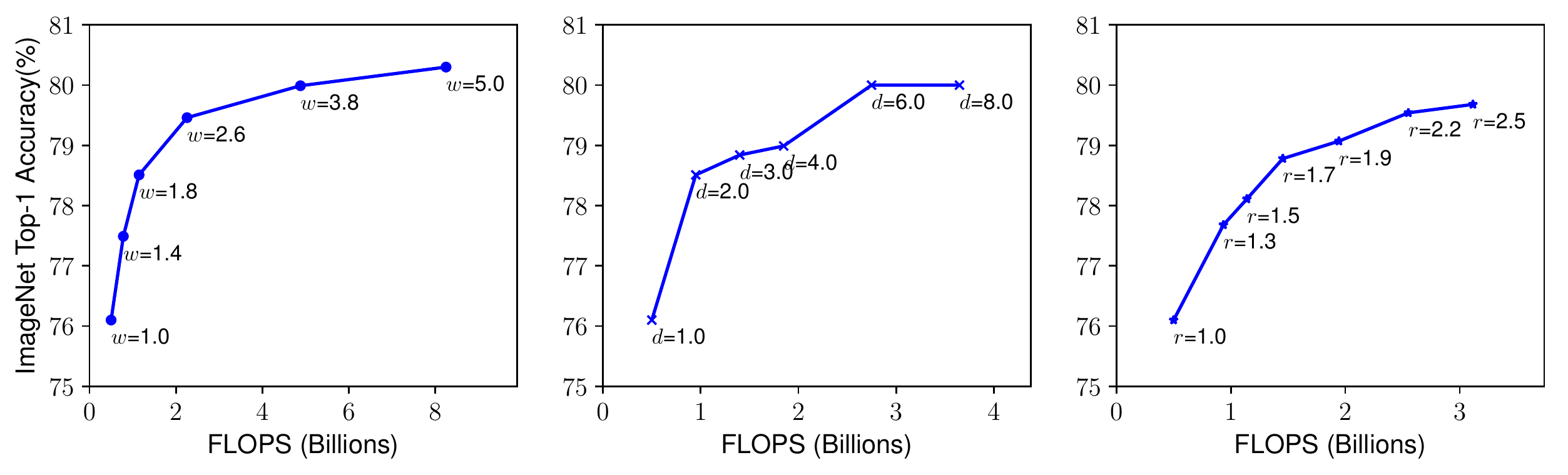}
        \vskip -0.1in
        \caption{\textbf{Scaling Up a Baseline Model with Different Network Width ($w$),  Depth ($d$), and Resolution ($r$) Coefficients.} Bigger networks with larger width, depth, or resolution tend to achieve higher accuracy, but the accuracy gain quickly saturate after reaching 80\%, demonstrating the limitation of single dimension scaling. Baseline network is described in Table \ref{tab:efficientnetb0}.
}
        \label{fig:scale-single}    

\end{figure*} 
\section{Compound Model Scaling}
\label{sec:study}
 
In this section, we will formulate the scaling problem,  study different approaches, and propose our new scaling method.

\subsection{Problem Formulation}
 
A ConvNet Layer $i$ can be defined as a function: $Y_i = \M{F}_i(X_i)$, where $\M{F}_i$ is the operator, $Y_i$ is output tensor, $X_i$ is input tensor, with tensor shape  $\langle H_i, W_i, C_i \rangle$\footnote{For the sake of simplicity, we omit batch dimension.}, where $H_i$ and $W_i$ are spatial dimension and $C_i$ is the channel dimension. A ConvNet $\M{N}$ can be represented by a list of composed layers:  $\M{N} = \M{F}_k \odot ... \odot  \M{F}_2 \odot \M{F}_1 (X_1) = \bigodot_{j=1...k} \M{F}_j (X_1)$.  In practice, ConvNet layers are often partitioned into multiple stages and all layers in each stage share the same architecture: for example,  ResNet \cite{resnet16} has five stages, and all layers in each stage has the same convolutional type except the first layer performs down-sampling. Therefore, we can define a ConvNet as:

\vspace{-0.1in}
\begin{equation}
\M{N} = \bigodot_{i=1...s} \M{F}_{i}^{L_i} \big(X_{\langle H_i, W_i, C_i \rangle}\big)
\end{equation}
\vspace{-0.1in}

where $\M{F}_{i}^{L_i}$ denotes layer $F_{i}$ is repeated $L_i$ times in stage $i$,  $\langle H_i, W_i, C_i \rangle$ denotes the shape of input tensor $X$ of layer $i$.   Figure \ref{fig:scalecompare}(a) illustrate a representative ConvNet, where the spatial dimension is gradually shrunk but the channel dimension is  expanded over layers, for example, from  initial input shape $\langle 224, 224, 3 \rangle$ to  final output shape $\langle 7, 7, 512 \rangle$.

Unlike regular ConvNet designs that mostly focus on finding the best layer architecture $\M{F}_i$, model scaling tries to expand the network length ($L_i$), width ($C_i$), and/or resolution ($H_i, W_i$) without changing $\M{F}_i$ predefined in the baseline network.  By fixing $\M{F}_i$, model scaling simplifies the design problem for new resource constraints, but it still remains a large design space to explore different $L_i, C_i, H_i, W_i$ for each layer.  In order to further reduce the design space, we restrict that all layers must be scaled uniformly with constant ratio. Our target is to maximize the model accuracy for any given resource constraints, which can be formulated as an optimization problem:

\vspace{-0.2in}
\begin{equation} \label{eq:opt} 
\begin{aligned}    
&\operatorname*{max}_{d, w, r} & &Accuracy \big(\M{N}(d, w, r)\big) \\
&s.t. & & \M{N}(d, w, r) = \bigodot_{i=1...s} \M{\hat F}_{i}^{d\cdot \hat  L_i} \big(X_{ \langle r\cdot  \hat H_i, r\cdot  \hat W_i, w\cdot  \hat C_i \rangle}\big) \\
& && \text{Memory}(\M{N}) \le \text{target\_memory}  \\
&&& \text{FLOPS}(\M{N}) \le \text{target\_flops}   \\
\end{aligned}    
\end{equation}
\vspace{-0.2in}

where $w, d, r$ are coefficients for scaling  network width, depth, and resolution; $\M{\hat F}_i, \hat L_i, \hat H_i, \hat W_i, \hat C_i$ are predefined parameters in baseline network (see Table \ref{tab:efficientnetb0} as an example).
 
\subsection{Scaling Dimensions}

The main difficulty of problem \ref{eq:opt} is that the optimal $d, w, r$ depend on each other and the values change under different resource constraints. Due to this difficulty, conventional methods mostly scale ConvNets in one of these dimensions:

\paragraph{Depth ($\pmb d$): }  Scaling network depth is the most common way used by many ConvNets \cite{resnet16,densenet17,googlenet14,inceptionv316}. The intuition is that deeper ConvNet can capture richer and more complex features, and generalize well on new tasks. However, deeper networks are also more difficult to train due to the vanishing gradient problem \cite{wideresnet16}.  Although several techniques, such as skip connections \cite{resnet16} and batch normalization \cite{batchnorm15}, alleviate the training problem, the accuracy gain of very deep network diminishes: for example, ResNet-1000 has similar accuracy as ResNet-101 even though it has much more layers.  Figure \ref{fig:scale-single} (middle) shows our empirical study on scaling a baseline model with different depth coefficient $d$, further suggesting the diminishing accuracy return for very deep ConvNets.

\paragraph{Width ($\pmb w$): } Scaling network width is commonly used for small size models \cite{mobilenetv117, mobilenetv218, mnas18}\footnote{In some literature, scaling number of channels is called ``depth multiplier", which means the same as our width coefficient $w$.}. As discussed in \cite{wideresnet16}, wider networks tend to be able to capture more fine-grained features and are easier to train. However, extremely wide but shallow networks tend to have difficulties in capturing higher level features. Our empirical results in Figure \ref{fig:scale-single} (left) show that the accuracy quickly saturates when networks become much wider with larger $w$.

\paragraph{Resolution ($\pmb r$): } With higher resolution input images, ConvNets can potentially capture more fine-grained patterns. Starting from 224x224 in early ConvNets, modern ConvNets tend to use 299x299 \cite{inceptionv316} or 331x331 \cite{nas_imagenet18} for better accuracy. Recently, GPipe \cite{gpipe18} achieves state-of-the-art ImageNet accuracy with 480x480 resolution. Higher resolutions, such as 600x600, are also widely used in object detection ConvNets \cite{maskrcnn17,retinanet17}. Figure \ref{fig:scale-single} (right) shows the results of scaling network resolutions, where indeed higher resolutions improve accuracy, but the accuracy gain diminishes for very high resolutions ($r=1.0$ denotes resolution 224x224 and $r=2.5$ denotes resolution 560x560).

The above analyses lead us to the first observation: 

\noindent \paragraph{Observation 1 -- } Scaling up any dimension of  network width, depth, or resolution improves accuracy, but the accuracy gain diminishes for bigger models.

\subsection{Compound Scaling}

We empirically observe that different scaling dimensions are not independent. Intuitively, for higher resolution images, we should increase network depth, such that the larger receptive fields can help capture similar features that include more pixels in bigger images. Correspondingly, we should also increase network width when resolution is higher,  in order to capture more fine-grained patterns with more pixels in high resolution images. These intuitions suggest that we need to coordinate and balance different scaling dimensions rather than conventional single-dimension scaling.

\begin{figure}                                                 
        \centering                                                                  
        \includegraphics[width=0.95\columnwidth]{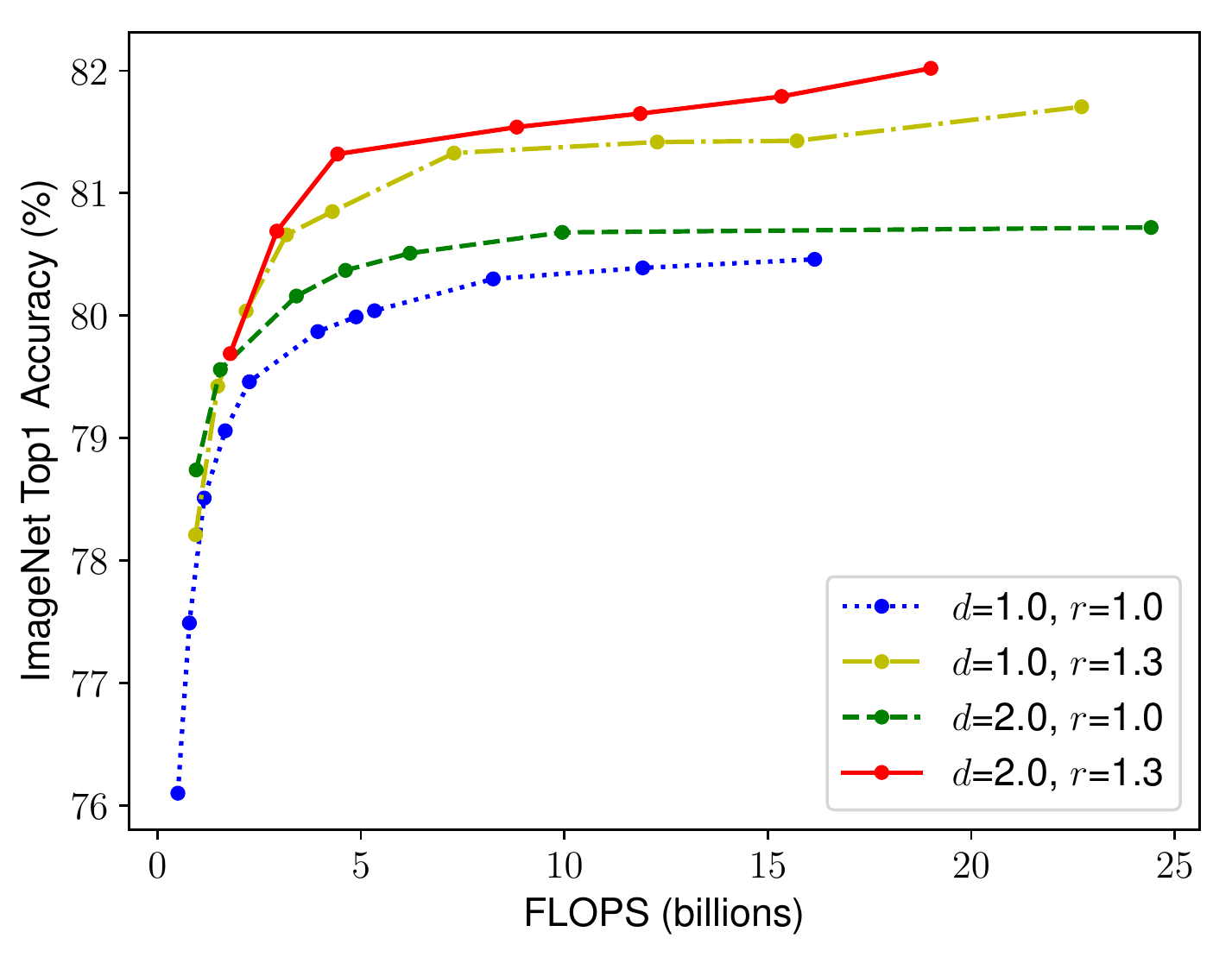}      
        \vskip -0.1in          
        \caption{\textbf{Scaling Network Width for Different Baseline Networks}. Each dot in a line denotes a model with different width coefficient ($w$). All baseline networks are from Table \ref{tab:efficientnetb0}.
        The first  baseline network ($d$=1.0, $r$=1.0) has 18 convolutional layers  with resolution 224x224, while the last baseline ($d$=2.0, $r$=1.3) has 36 layers with resolution 299x299. 
}                                                                           
        \label{fig:scale-combine}  
        \vskip  -0.2in 

\end{figure} 
To validate our intuitions, we compare width scaling under different network depths and resolutions, as shown in Figure \ref{fig:scale-combine}.  If we only scale network width $w$ without changing depth ($d$=1.0) and resolution ($r$=1.0), the accuracy saturates quickly.  With deeper ($d$=2.0) and higher resolution ($r$=2.0), width scaling  achieves much better accuracy under the same FLOPS cost. These results lead us to the second observation:

\noindent \paragraph{Observation 2 -- } In order to pursue better accuracy and efficiency, it is critical to balance all dimensions of network width, depth, and resolution during ConvNet scaling.

In fact, a few prior work \cite{nas_imagenet18, amoebanets18} have already tried to arbitrarily balance network width and depth, but they all require tedious manual tuning. 
 
In this paper, we propose a new \textbf{compound scaling method}, which use a compound  coefficient $\phi$ to 
uniformly scales network width, depth, and resolution in a  principled way:

\vspace{-0.1in}
\begin{equation} \label{eq:optobj} 
\begin{aligned}
\text{depth: } & d  = \alpha ^ \phi \\
\text{width: } & w = \beta ^ \phi \\
\text{resolution: } & r   =  \gamma ^ \phi  \\
\text{s.t.    }  & \alpha \cdot \beta ^2 \cdot \gamma ^ 2 \approx 2 \\
& \alpha \ge 1, \beta \ge 1, \gamma \ge 1 
\end{aligned}
\end{equation}

where $\alpha, \beta, \gamma$ are constants that can be determined by a small grid search.  Intuitively,  $\phi$ is a user-specified coefficient that controls how many more resources are available for model scaling, while $\alpha, \beta, \gamma$  specify how to assign these extra resources to network width, depth, and resolution respectively. Notably, the FLOPS of a regular convolution op is proportional to $d$, $w^2$, $r^2$, i.e., doubling network depth will double FLOPS, but doubling network width or resolution will increase FLOPS by four times. Since convolution ops usually dominate the computation cost in ConvNets, scaling a ConvNet with equation \ref{eq:optobj} will approximately increase total FLOPS by $\big(\alpha \cdot \beta ^2 \cdot \gamma ^2 \big)^\phi$.
In this paper, we constraint $\alpha \cdot \beta ^2 \cdot \gamma ^ 2 \approx 2$ such that for any new $\phi$, the total FLOPS will approximately\footnote{FLOPS may differ from theoretical value due to rounding.} increase by $2^\phi$. %

\section{EfficientNet Architecture}
\label{sec:method}

Since model scaling does not change layer operators $\M{\hat F}_i$ in baseline network, having a good baseline network is also critical.  We will evaluate our scaling method using existing ConvNets, but in order to better demonstrate the effectiveness of our scaling method, we have also developed a new mobile-size baseline, called EfficientNet.

Inspired by \cite{mnas18}, we develop our baseline network by leveraging a multi-objective neural architecture search that optimizes both accuracy and FLOPS.
Specifically, we use the same search space as ~\cite{mnas18},  and use $ACC(m) \times [FLOPS(m)/T]^w$ as the optimization goal, where $ACC(m)$ and $FLOPS(m)$ denote the accuracy and FLOPS of model $m$, $T$ is the target FLOPS and $w$=-0.07 is a hyperparameter for controlling the trade-off between accuracy and FLOPS. Unlike \cite{mnas18,proxyless18}, here we optimize FLOPS rather than latency since we are not targeting any specific hardware device. Our search produces an efficient network, which we name EfficientNet-B0. Since we use the same search space as~\cite{mnas18}, the architecture is similar to MnasNet, except our EfficientNet-B0 is slightly bigger due to the larger FLOPS target (our FLOPS target is 400M).
Table \ref{tab:efficientnetb0} shows the architecture of EfficientNet-B0. Its main building block is mobile inverted bottleneck MBConv \cite{mobilenetv218,mnas18}, to which we also add squeeze-and-excitation optimization \cite{senet18}.

\begin{table}
	\vskip -0.1in     
	\caption{	\BF{EfficientNet-B0 baseline network }--  Each row describes a stage $i$ with $\hat L_i$ layers, with input resolution $\langle \hat H_i, \hat W_i \rangle$ and output channels $\hat C_i$. Notations are adopted from equation \ref{eq:opt}.
	}
  \vskip 0.05in
  \centering   

  \resizebox{1.0\columnwidth}{!}{ 
	\begin{tabular}{c|c|c|c|c}                                                     
	\toprule[0.15em]                                                                  
	Stage & Operator  & Resolution      & \#Channels & \#Layers \\
	$i$ &   $\mathcal{ \hat F}_i$ & $ \hat H_i \times  \hat W_i $     &  $ \hat  C_i$ & $ \hat L_i$        \\
	\midrule                                                              
	1  &    Conv3x3          & $ 224 \times 224 $   & 32 & 1 \\
	2 &   MBConv1, k3x3   & $112 \times 112 $  & 16   &1 \\
	3 &   MBConv6, k3x3    & $112 \times 112$ & 24   & 2 \\
	4 &   MBConv6, k5x5   &  $56 \times 56$ &  40   & 2  \\
	5 &   MBConv6, k3x3    & $28 \times 28 $     & 80   &  3 \\
	6 &   MBConv6, k5x5  & $14 \times 14$  &  112  & 3 \\
	7 &   MBConv6, k5x5  & $14 \times 14$ &   192  & 4  \\
	8 & MBConv6, k3x3   & $7 \times 7$ &   320  & 1  \\
	9 & Conv1x1 \& Pooling \& FC   & $7 \times 7$ & 1280 & 1 \\       
	\bottomrule[0.15em]                                                                     
\end{tabular}                                                                  
  }                                                                                 
              
  \label{tab:efficientnetb0}      
\end{table}                     %

Starting from the baseline EfficientNet-B0, we apply our compound scaling method to scale it up with two steps:

\begin{itemize}
	\setlength\itemsep{0em}
	\item STEP 1: we first fix $\phi=1$, assuming twice more resources available, and do a small grid search of $\alpha, \beta,\gamma$ based on Equation \ref{eq:opt} and \ref{eq:optobj}. In particular, we find the best values for EfficientNet-B0 are $\alpha=1.2, \beta=1.1, \gamma=1.15$, under constraint of $\alpha \cdot \beta ^2 \cdot \gamma^2 \approx 2$. 
	\item STEP 2: we then fix $\alpha, \beta,\gamma$ as constants and scale up baseline network with different $\phi$ using Equation \ref{eq:optobj}, to obtain EfficientNet-B1 to B7 (Details in Table \ref{tab:imagenet}).
\end{itemize}

Notably, it is possible to achieve even better performance by searching for  $\alpha, \beta,\gamma$  directly around a large model, but the search cost becomes prohibitively more expensive on larger models. Our method solves this issue by only doing search once on the small baseline network (step 1), and then use the same  scaling coefficients for all other models (step 2).

\begin{table*}
    \caption{
        \textbf{EfficientNet Performance Results on ImageNet} \cite{imagenet15}. All EfficientNet models are scaled from our baseline EfficientNet-B0 using different compound coefficient $\phi$ in Equation \ref{eq:optobj}.
        ConvNets with similar top-1/top-5 accuracy are grouped together for efficiency comparison. Our scaled EfficientNet models consistently reduce parameters and FLOPS by an order of magnitude (up to 8.4x parameter reduction and up to 16x FLOPS reduction) than existing ConvNets.
       }
  \vskip 0.1in
    \centering
    \resizebox{1.0\textwidth}{!}{
        \begin{tabular}{l|cc||cc||cc}
        \toprule [0.15em]
        Model &  Top-1 Acc. & Top-5 Acc. & \#Params &  Ratio-to-EfficientNet &\#FLOPs & Ratio-to-EfficientNet  \\
        \midrule [0.1em]
        \bf EfficientNet-B0  & \bf 77.1\% & \bf 93.3\% & \bf 5.3M  & \bf 1x & \bf 0.39B & \bf  1x \\
        ResNet-50 \cite{resnet16} & 76.0\% & 93.0\% & 26M & 4.9x & 4.1B & 11x \\
        DenseNet-169 \cite{densenet17} & 76.2\% & 93.2\% & 14M & 2.6x & 3.5B & 8.9x \\

        \midrule
        \bf EfficientNet-B1  & \bf 79.1\% & \bf 94.4\% & \bf 7.8M  & \bf 1x & \bf 0.70B & \bf  1x \\
        ResNet-152 \cite{resnet16} & 77.8\% & 93.8\% & 60M & 7.6x & 11B & 16x \\
        DenseNet-264 \cite{densenet17} & 77.9\% & 93.9\% & 34M & 4.3x & 6.0B & 8.6x \\
        Inception-v3 \cite{inceptionv316} & 78.8\% & 94.4\% & 24M & 3.0x & 5.7B & 8.1x \\
        Xception \cite{xception17} & 79.0\% & 94.5\% & 23M & 3.0x & 8.4B & 12x \\

        \midrule
        \bf EfficientNet-B2  & \bf 80.1\% & \bf 94.9\% & \bf 9.2M  & \bf 1x & \bf 1.0B & \bf  1x \\
        Inception-v4 \cite{inceptionv417} & 80.0\% & 95.0\% & 48M & 5.2x & 13B & 13x \\
        Inception-resnet-v2  \cite{inceptionv417} & 80.1\% & 95.1\% & 56M & 6.1x & 13B & 13x \\

         \midrule
		\bf EfficientNet-B3  & \bf 81.6\% & \bf 95.7\% & \bf 12M  & \bf 1x & \bf 1.8B & \bf  1x \\
		ResNeXt-101 \cite{resnext17} & 80.9\% & 95.6\% & 84M & 7.0x & 32B & 18x \\
		PolyNet \cite{polynet17} & 81.3\%  & 95.8\% & 92M & 7.7x & 35B & 19x \\

		\midrule
		\bf EfficientNet-B4  & \bf 82.9\% & \bf 96.4\% & \bf 19M  & \bf 1x & \bf 4.2B & \bf  1x \\
		SENet \cite{senet18} & 82.7\% & 96.2\% & 146M & 7.7x & 42B & 10x \\
		 NASNet-A \cite{nas_imagenet18} & 82.7\% & 96.2\% & 89M & 4.7x & 24B & 5.7x \\
		 AmoebaNet-A \cite{amoebanets18} & 82.8\% & 96.1\% & 87M & 4.6x & 23B & 5.5x \\
		 PNASNet \cite{pnas18} & 82.9\% & 96.2\% & 86M & 4.5x & 23B & 6.0x \\

		\midrule
		\bf EfficientNet-B5  & \bf 83.6\% & \bf 96.7\% & \bf 30M  & \bf 1x & \bf 9.9B & \bf  1x \\
		 AmoebaNet-C  \cite{autoaugment18} &  83.5\%  & 96.5\% & 155M & 5.2x & 41B & 4.1x \\

		\midrule
		 \bf EfficientNet-B6  & \bf 84.0\% & \bf 96.8\% & \bf 43M  & \bf 1x & \bf 19B & \bf  1x \\
		\midrule
		\bf EfficientNet-B7  & \bf 84.3\% & \bf 97.0\% & \bf 66M  & \bf 1x & \bf 37B & \bf  1x \\
		GPipe \cite{gpipe18}  & 84.3\% & 97.0\% & 557M & 8.4x & - & - \\
        \bottomrule[0.15em]
        \multicolumn{7}{l}{~~We omit ensemble and multi-crop models \cite{senet18}, or models pretrained on 3.5B Instagram images \cite{imagenetinstagram18}.~~}
        \end{tabular}
    }
    \label{tab:imagenet}
\end{table*}
 
\section{Experiments}
\label{sec:results}

\begin{table}
  \vskip -0.2in
  \caption{ 
      \textbf{Scaling Up MobileNets and ResNet.}
  }                                                                                 
  \vskip 0.05in
  \centering   
  \resizebox{0.98\columnwidth}{!}{ 
  	\begin{tabular}{l|cc}                
  		  \toprule[0.15em]                                                 
         Model      &   FLOPS  & Top-1 Acc.  \\             
		  \midrule[0.12em]                                                 
          Baseline MobileNetV1  \cite{mobilenetv117}     & 0.6B  &  70.6\% \\
          \midrule[0.05em]
	      Scale MobileNetV1 by width ($w$=2)     & 2.2B   &  74.2\% \\
		  Scale MobileNetV1 by resolution ($r$=2)&2.2B &  72.7\%  \\
		 \bf compound scale ($\pmb d$=1.4, $\pmb w$=1.2, $\pmb r$=1.3)      & \bf 2.3B &  \bf 75.6\% \\
		  \midrule[0.15em]                                                 
			Baseline MobileNetV2 \cite{mobilenetv218}  & 0.3B  &  72.0\%  \\
			\midrule[0.05em]
			Scale MobileNetV2 by depth ($d$=4)     & 1.2B   &  76.8\% \\
			Scale MobileNetV2 by width ($w$=2)    & 1.1B   &  76.4\%  \\
			Scale MobileNetV2 by resolution ($r$=2)   & 1.2B &  74.8\% \\
			\bf MobileNetV2 compound scale &  \bf 1.3B &  \bf 77.4\% \\ 

            \midrule[0.15em]
			Baseline ResNet-50 \cite{resnet16}  & 4.1B  &  76.0\%  \\
			\midrule[0.05em]
			Scale ResNet-50  by depth ($d$=4)     & 16.2B   &  78.1\% \\
			Scale ResNet-50 by width ($w$=2)    & 14.7B   &  77.7\%  \\
			Scale ResNet-50 by resolution ($r$=2)   & 16.4B &  77.5\% \\
			\bf ResNet-50 compound scale &  \bf 16.7B &  \bf 78.8\% \\ 

          \bottomrule[0.15em]
    \end{tabular}                                                                 
  }                                                                                 
  \label{tab:mobilenets}
\end{table} 
 
In this section, we will first evaluate our scaling method on existing ConvNets and the new proposed EfficientNets.

\subsection{Scaling Up MobileNets and ResNets}

As a proof of concept, we first apply our scaling method to the widely-used MobileNets \cite{mobilenetv117,mobilenetv218} and ResNet \cite{resnet16}. Table \ref{tab:mobilenets} shows the ImageNet results of scaling them in different ways. Compared to other single-dimension scaling methods, our compound scaling method improves the accuracy on all these models, suggesting the effectiveness of our proposed scaling method for general existing ConvNets.

\begin{table}
 \vskip -0.1in
  \caption{                                                                         
	\textbf{Inference Latency Comparison} -- Latency is measured with batch size 1 on a single core of Intel Xeon CPU E5-2690.
}          
  \vskip 0.1in
  \centering                                                                        
  \resizebox{0.99\columnwidth}{!}{ 
  	\begin{tabular}{cc||cc}      
  		  \toprule[0.15em]                                                           
               & Acc. @ Latency  &  & Acc. @ Latency \\
          \midrule[0.1em]
          ResNet-152 & 77.8\% @ 0.554s & GPipe &  84.3\% @ 19.0s  \\
          EfficientNet-B1& 78.8\% @ 0.098s & EfficientNet-B7 &  84.4\%  @ 3.1s  \\
          \bf Speedup &  \bf 5.7x  & \bf Speedup & \bf 6.1x \\
          \bottomrule[0.15em]                                                           
    \end{tabular}                                                                 
  }                                                                                 
                                                                       
  \label{tab:latency}      
  \vskip -0.1in
\end{table} 
\begin{figure}[!t]
	\includegraphics[width=0.96\columnwidth]{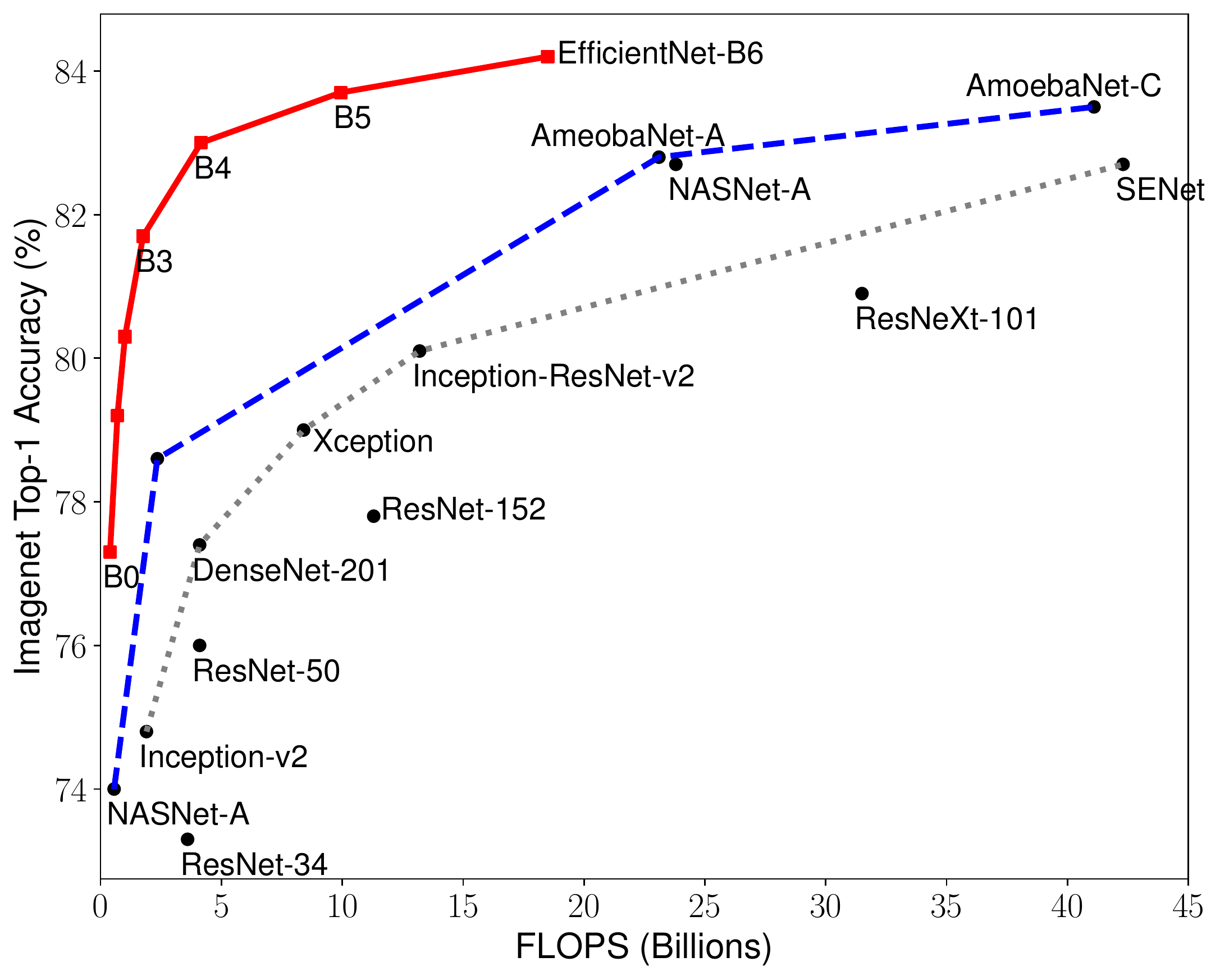}
	\hspace{-46mm}\resizebox{.5\columnwidth}{!}{\tablestyle{2pt}{1}
		\begin{tabular}[b]{l|cr}
			& Top1 Acc. & FLOPS  \\
			\shline
			ResNet-152 \cite{resnext17}         &  77.8\% & 11B \\
			\bf EfficientNet-B1  & \bf 79.1\%  & \bf 0.7B \\
			\hline
			ResNeXt-101 \cite{resnext17}         &  80.9\% & 32B \\
			\bf EfficientNet-B3                                    & \bf 81.6\%  & \bf 1.8B \\
			\hline
			SENet  \cite{senet18}                       &  82.7\% & 42B \\
			NASNet-A \cite{nas_imagenet18}	& 80.7\% & 24B \\            %
            \bf EfficientNet-B4                                    & \bf 82.9\% & \bf 4.2B \\
			\hline
			AmeobaNet-C \cite{autoaugment18}  &  83.5\% & 41B \\
			\bf EfficientNet-B5                                    & \bf 83.6\% & \bf 9.9B \\

			\multicolumn{3}{l}{\vspace{12mm} } \\
	\end{tabular}}
	\vskip -0.15in
	\caption{\BF{FLOPS vs. ImageNet Accuracy --} Similar to Figure \ref{fig:imagnet-params} except it compares FLOPS rather than model size.}
	\label{fig:imagenet-flops}
	\vskip -0.1in
\end{figure}

\begin{table*}      
	    \caption{                                                                       
		\textbf{EfficientNet Performance Results on Transfer Learning Datasets}. Our scaled EfficientNet models achieve new state-of-the-art accuracy for 5 out of 8 datasets, with 9.6x fewer parameters on average.
	}         
   \vskip 0.1in
    \centering                                                                      
    \resizebox{1.0\textwidth}{!}{                                                  
        \begin{tabular}{l||cccccc||cccccc}                                                
        \toprule [0.15em]
              &   \multicolumn{6}{c}{Comparison to best public-available results} & \multicolumn{6}{c}{Comparison to best reported results} \\
              &  Model & Acc. & \#Param &  Our Model & Acc. & \#Param(ratio) &  Model & Acc. & \#Param & Our Model & Acc. & \#Param(ratio)   \\
        \midrule [0.1em]
CIFAR-10 &  NASNet-A  & 98.0\% & 85M & EfficientNet-B0 & 98.1\% &  4M (21x) & $^\dagger$Gpipe  &  \bf 99.0\% & 556M & EfficientNet-B7 & 98.9\% &  64M (8.7x) \\

CIFAR-100 &  NASNet-A & 87.5\% & 85M & EfficientNet-B0 & 88.1\% &  4M (21x) & Gpipe & 91.3\% & 556M & EfficientNet-B7 &  \bf  91.7\% &  64M (8.7x) \\

Birdsnap  &  Inception-v4 & 81.8\% & 41M & EfficientNet-B5 & 82.0\% &  28M (1.5x) & GPipe & 83.6\% & 556M & EfficientNet-B7 &  \bf  84.3\% &   64M (8.7x) \\

Stanford Cars  & Inception-v4  & 93.4\% & 41M & EfficientNet-B3 & 93.6\% &  10M (4.1x) & $^\ddagger$DAT &   \bf 94.8\%& - & EfficientNet-B7 & 94.7\% &  - \\

Flowers  &  Inception-v4 & 98.5\% & 41M & EfficientNet-B5 & 98.5\% &  28M (1.5x) & DAT & 97.7\% & - & EfficientNet-B7 &  \bf 98.8\% &  - \\

FGVC Aircraft  &  Inception-v4  & 90.9\% & 41M & EfficientNet-B3 &  90.7\% &  10M (4.1x) & DAT  & 92.9\% & - & EfficientNet-B7 &  \bf  92.9\% & -  \\

Oxford-IIIT Pets  &  ResNet-152 & 94.5\% & 58M & EfficientNet-B4 & 94.8\% &  17M (5.6x) & GPipe & \bf  95.9\% & 556M & EfficientNet-B6 & 95.4\% &  41M (14x) \\

Food-101  &  Inception-v4 & 90.8\% & 41M & EfficientNet-B4 & 91.5\% &  17M (2.4x) & GPipe  & 93.0\% & 556M & EfficientNet-B7 &  \bf  93.0\% &  64M (8.7x) \\
\midrule
        Geo-Mean & & & & & &  \bf (4.7x) &  & & & & &  \bf (9.6x)\\
        \bottomrule[0.15em]
        \multicolumn{13}{l}{~~$^\dagger$GPipe \cite{gpipe18} trains giant models with specialized pipeline parallelism library.} \\
        \multicolumn{13}{l}{~~$^\ddagger$DAT denotes domain adaptive transfer learning \cite{domainjft18}. Here we only compare ImageNet-based transfer learning results.~~} \\
        \multicolumn{13}{l}{~~Transfer accuracy and \#params for NASNet \cite{nas_imagenet18}, Inception-v4 \cite{inceptionv417}, ResNet-152 \cite{resnet16} are from \cite{imagenettransfer18}.~~} \\
        \end{tabular}                                                            
    }                                                                   
    \label{tab:transfer}                                                         
\end{table*} %
\begin{figure*}
        \centering
        \includegraphics[width=0.8\textwidth]{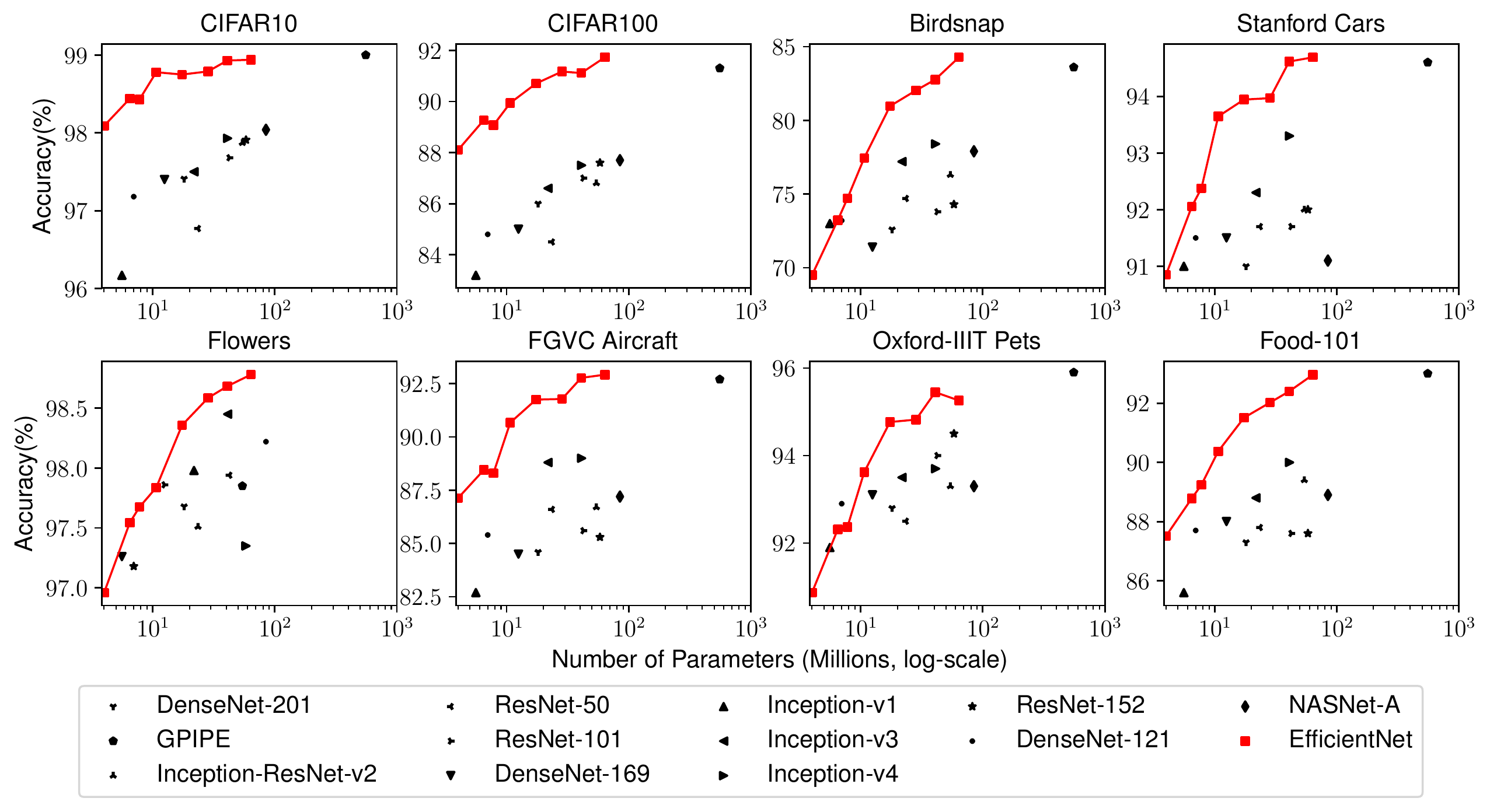}
         \vskip -0.2in
        \caption{\textbf{Model Parameters vs. Transfer Learning Accuracy -- } All models are pretrained on ImageNet and finetuned on new datasets. }
        \label{fig:transfer-all}
         \vskip -0.1in
\end{figure*} 
\subsection{ImageNet Results for EfficientNet}

We train our EfficientNet models on ImageNet using similar settings as \cite{mnas18}: RMSProp optimizer with decay 0.9 and momentum 0.9; batch norm  momentum 0.99; weight decay 1e-5;  initial learning rate 0.256 that decays by 0.97 every 2.4 epochs. We also use SiLU (Swish-1) activation \cite{swish18,swishsil18,gelu16},  AutoAugment \cite{autoaugment18}, and stochastic depth \cite{droppath16} with survival probability 0.8. As commonly known that bigger models need more regularization, we linearly increase dropout \cite{dropout14} ratio from 0.2 for EfficientNet-B0 to 0.5 for B7. We reserve 25K randomly picked images from the \TT{training} set as a \TT{minival} set, and perform early stopping on this \TT{minival}; we then evaluate the early-stopped checkpoint on the original \TT{validation} set to report the final validation accuracy.

Table \ref{tab:imagenet} shows the performance of all EfficientNet models that are scaled from the same baseline EfficientNet-B0. Our EfficientNet models generally use  an order of magnitude fewer parameters and FLOPS than other ConvNets with similar accuracy. In particular, our EfficientNet-B7 achieves 84.3\% top1 accuracy with 66M parameters and 37B FLOPS, being more accurate but \BF{8.4x smaller} than the previous best GPipe \cite{gpipe18}.  These gains come from both better architectures, better scaling, and better training settings that are customized for EfficientNet.

Figure \ref{fig:imagnet-params} and Figure \ref{fig:imagenet-flops} illustrates the parameters-accuracy and FLOPS-accuracy curve for representative ConvNets, where our scaled EfficientNet models achieve better accuracy with much fewer parameters and FLOPS than other ConvNets. Notably, our EfficientNet models are not only small, but also computational cheaper. For example, our EfficientNet-B3 achieves higher accuracy than ResNeXt-101 \cite{resnext17} using \BF{18x fewer FLOPS}.

To validate the latency, we have also measured the inference latency for a few representative CovNets on a real CPU as shown in Table \ref{tab:latency}, where we report average latency over 20 runs. Our EfficientNet-B1 runs  \BF{5.7x faster} than the widely used ResNet-152, while EfficientNet-B7 runs about \BF{6.1x faster} than GPipe \cite{gpipe18}, suggesting our EfficientNets are indeed fast on real hardware.

\begin{figure*}                                                          
        \centering                                                                  
        \includegraphics[width=0.9\textwidth]{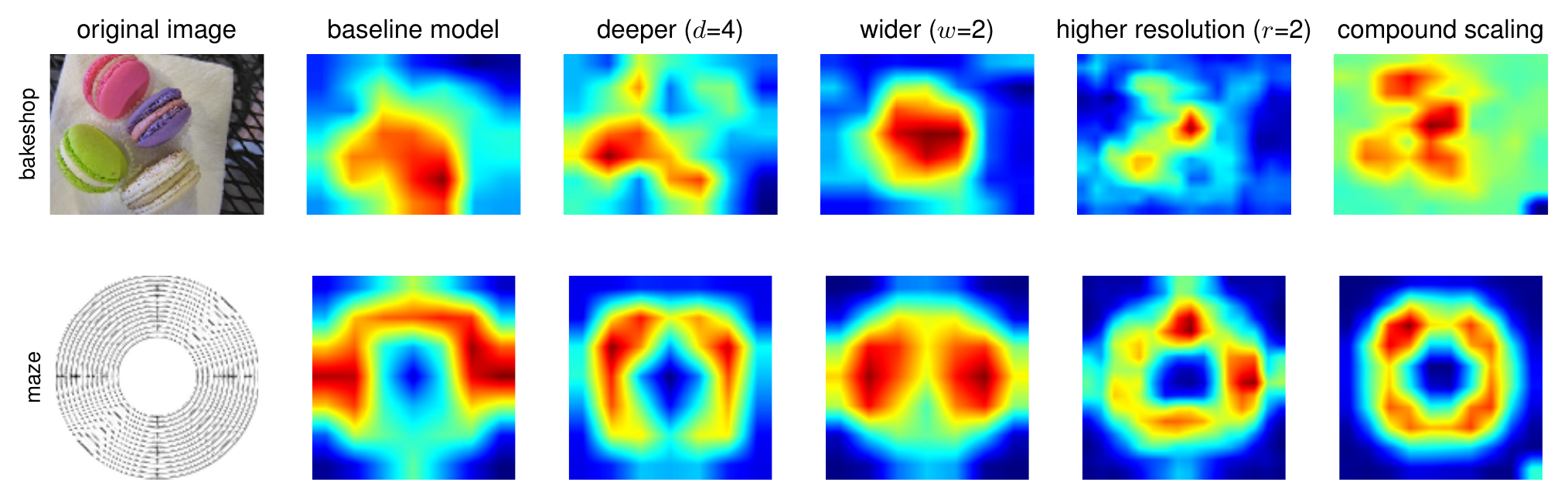} 
        \vskip -0.1in
        \caption{\textbf{Class Activation Map (CAM) \cite{cam16} for Models with different scaling methods-} Our compound scaling method allows the scaled model (last column) to focus on more relevant regions with more object details. Model details are in Table \ref{tab:cam-model}.
         }                                                                           
        \label{fig:cam}
         \vskip -0.15in  
\end{figure*} 

\subsection{Transfer Learning Results for EfficientNet}
\begin{table}                                                
  \caption{                                                                         
      \textbf{Transfer Learning Datasets}.  
  }                                                                                 
  \vskip 0.1in
  \centering   
  \resizebox{1.0\columnwidth}{!}{ 
  	\begin{tabular}{c|ccc}         
  		\toprule[0.15em]                                                        
         Dataset      & Train Size  &   Test Size & \#Classes \\             
         \midrule[0.1em]
          CIFAR-10 \cite{cifar} &  50,000 & 10,000 & 10 \\
          CIFAR-100 \cite{cifar} &  50,000 & 10,000 & 100  \\
	      Birdsnap \cite{birdsnap} &  47,386 & 2,443 & 500 \\
		 Stanford Cars \cite{stanfordcars} &  8,144 & 8,041 & 196 \\
		 Flowers \cite{flowers} & 2,040 & 6,149 & 102 \\
		 FGVC Aircraft \cite{aircraft} &  6,667 & 3,333 & 100 \\
		Oxford-IIIT Pets \cite{oxfordpets} & 3,680 & 3,369 & 37 \\
		Food-101 \cite{food101} & 75,750 & 25,250 & 101   \\
         \bottomrule[0.1em]                                                           
    \end{tabular}                                                                 
  }                                                                                 
  \label{tab:transfer-dataset}
  \vskip -0.15in
\end{table} 
 
We have also evaluated our EfficientNet on a list of commonly used transfer learning datasets, as shown in Table \ref{tab:transfer-dataset}. We borrow the same training settings from \cite{imagenettransfer18} and \cite{gpipe18}, which take ImageNet pretrained checkpoints and finetune on new datasets.

Table \ref{tab:transfer} shows the transfer learning performance: (1) Compared to public available models, such as NASNet-A \cite{nas_imagenet18} and Inception-v4 \cite{inceptionv417}, our EfficientNet models achieve better accuracy with 4.7x average (up to 21x) parameter reduction.  (2) Compared to state-of-the-art models, including DAT \cite{domainjft18} that dynamically synthesizes training data and GPipe \cite{gpipe18} that is trained with specialized pipeline parallelism, our EfficientNet models still surpass their accuracy in 5 out of 8 datasets, but using 9.6x fewer parameters

Figure \ref{fig:transfer-all} compares the accuracy-parameters curve for a variety of models. In general, our EfficientNets  consistently achieve better accuracy with an order of magnitude fewer parameters than existing models, including ResNet \cite{resnet16}, DenseNet \cite{densenet17}, Inception \cite{inceptionv417}, and NASNet \cite{nas_imagenet18}.

\section{Discussion}
\label{sec:discuss}
\begin{figure}[!t]
	\centering
	\includegraphics[width=0.85\columnwidth]{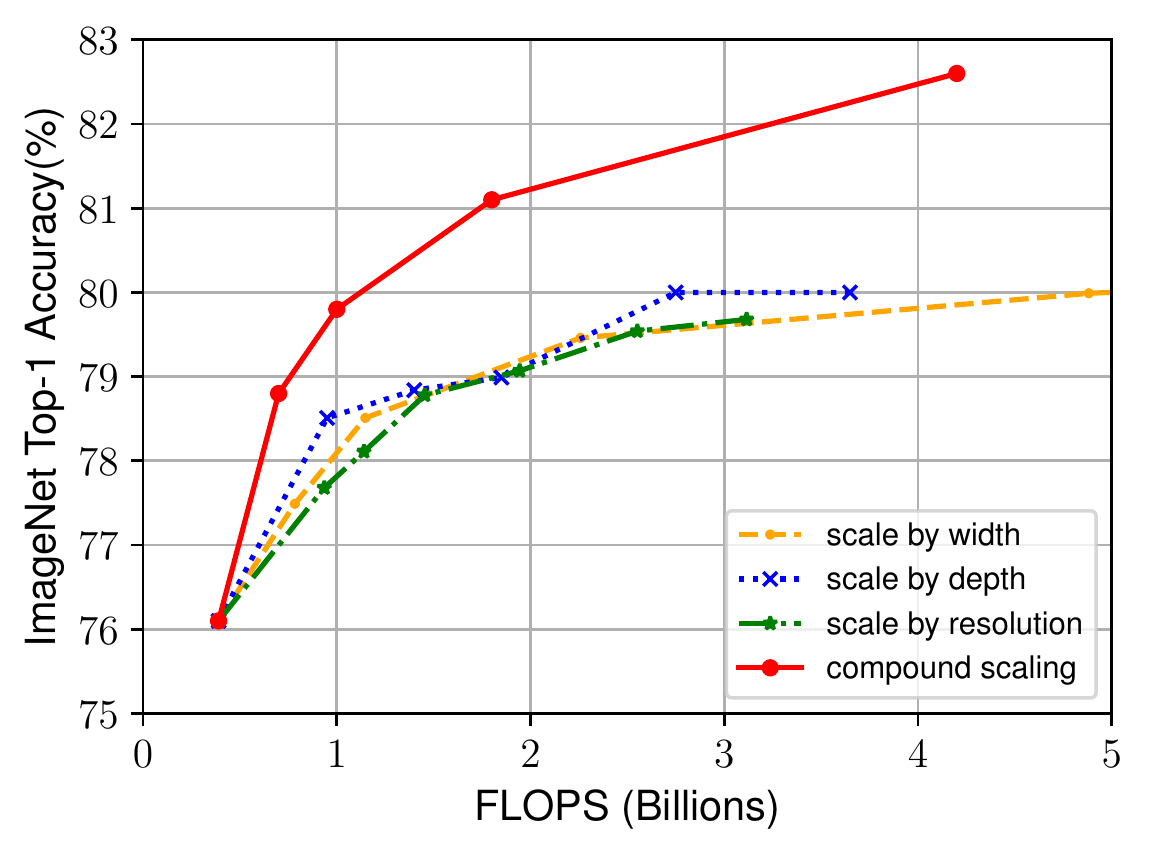}
	\vskip -0.2in
	\caption{\BF{Scaling Up EfficientNet-B0 with Different Methods.}
	}
	\label{fig:scale-flopscompare}
	\vskip -0.1in
\end{figure}
\begin{table}                                                               
  \caption{                                                                         
      \textbf{Scaled Models Used in Figure \ref{fig:cam}.}
  }                                                                                 
  \vskip 0.1in
  \centering   

  \resizebox{0.95\columnwidth}{!}{ 
  	\begin{tabular}{l|cc}                
  		  \toprule[0.15em]                                                 
         Model      &    FLOPS  & Top-1 Acc.  \\             
		  \midrule[0.1em]                                                 
          Baseline model (EfficientNet-B0)          & 0.4B  &  77.3\%  \\
          \midrule
          Scale model by depth ($d$=4)     & 1.8B   &  79.0\%  \\
	      Scale model by width ($w$=2)     & 1.8B   &  78.9\%  \\
		  Scale model by resolution ($r$=2)  & 1.9B &  79.1\%  \\
		  \bf Compound Scale ($\pmb d$=1.4, $\pmb w$=1.2, $\pmb r$=1.3) &  \bf 1.8B &  \bf 81.1\% \\ 
          \bottomrule[0.15em]                                                           
    \end{tabular}                                                                 
  }                                                                                 
  \label{tab:cam-model}
  \vskip -0.1in
\end{table}

To disentangle the contribution of our proposed scaling method from the EfficientNet architecture, Figure \ref{fig:scale-flopscompare} compares the ImageNet performance of different scaling methods for the same EfficientNet-B0 baseline network.  In general, all scaling methods improve accuracy with the cost of more FLOPS, but our compound scaling method can further improve accuracy, by up to 2.5\%, than other single-dimension scaling methods, suggesting the importance of our proposed compound scaling.

In order to further understand why our compound scaling method is better than others, Figure \ref{fig:cam} compares the class activation map \cite{cam16} for a few representative models with different scaling methods. All these models are scaled from the same baseline, and their statistics are shown in Table \ref{tab:cam-model}. Images are randomly picked from ImageNet validation set.  As shown in the figure, the model with compound scaling tends to focus on more relevant regions with more object details, while other models are either lack of object details or unable to capture all objects in the images.
\section{Conclusion}
\label{sec:conclusion}

In this paper, we systematically study ConvNet scaling and identify that carefully balancing network width, depth, and resolution is an important but missing piece, preventing us from better accuracy and efficiency. To address this issue, we propose a simple and highly effective compound scaling method, which enables us to easily scale up a baseline ConvNet to any target resource constraints in a more principled way, while maintaining model efficiency.  Powered by this compound scaling method, we demonstrate that a mobile-size EfficientNet model can be scaled up very effectively, surpassing state-of-the-art accuracy with an order of magnitude fewer parameters and FLOPS, on both ImageNet and five commonly used  transfer learning datasets.

\section*{Acknowledgements}
We thank Ruoming Pang, Vijay Vasudevan, Alok Aggarwal, Barret Zoph, Hongkun Yu, Jonathon Shlens, Raphael Gontijo Lopes, 
Yifeng Lu, Daiyi Peng, Xiaodan Song, Samy Bengio, Jeff Dean, and the Google Brain team for their help.

\section*{Appendix}
Since 2017, most research papers only report and compare ImageNet validation accuracy; this paper also follows this convention for better comparison. In addition, we have also verified the test accuracy by submitting our predictions on the 100k test set images to {\small\url{http://image-net.org}}; results are in Table \ref{tab:testacc}. As expected, the test accuracy is very close to the validation accuracy.

\begin{table}[!h]
    \vskip -0.1in
    \caption{\textbf{ImageNet Validation vs. Test Top-1/5 Accuracy}. }
     \centering
     \resizebox{0.99\columnwidth}{!}{
         \begin{tabular}{r|cccccccc}
             \toprule[0.15em]
                  & B0 & B1 & B2 & B3 & B4 & B5 & B6 & B7 \\
             \midrule[0.1em]
             Val top1       & 77.11 & 79.13 & 80.07 & 81.59 & 82.89 & 83.60 & 83.95 & 84.26  \\
             Test top1      & 77.23 & 79.17 & 80.16 & 81.72 & 82.94 & 83.69 & 84.04 & 84.33  \\
             \midrule[0.1em]
             Val top5       & 93.35 & 94.47 & 94.90 & 95.67 & 96.37 & 96.71 & 96.76 & 96.97 \\
             Test top5      & 93.45 & 94.43 & 94.98 & 95.70 & 96.27 & 96.64 & 96.86 & 96.94 \\
             \bottomrule[0.1em]
       \end{tabular}
     }

     \label{tab:testacc}
     \vskip -0.1in
   \end{table}

\bibliographystyle{sty/icml2019}
\bibliography{cv}

\begin{thebibliography}{52}
\providecommand{\natexlab}[1]{#1}
\providecommand{\url}[1]{\texttt{#1}}
\expandafter\ifx\csname urlstyle\endcsname\relax
  \providecommand{\doi}[1]{doi: #1}\else
  \providecommand{\doi}{doi: \begingroup \urlstyle{rm}\Url}\fi

\bibitem[Berg et~al.(2014)Berg, Liu, Woo~Lee, Alexander, Jacobs, and
  Belhumeur]{birdsnap}
Berg, T., Liu, J., Woo~Lee, S., Alexander, M.~L., Jacobs, D.~W., and Belhumeur,
  P.~N.
\newblock Birdsnap: Large-scale fine-grained visual categorization of birds.
\newblock \emph{CVPR}, pp.\  2011--2018, 2014.

\bibitem[Bossard et~al.(2014)Bossard, Guillaumin, and Van~Gool]{food101}
Bossard, L., Guillaumin, M., and Van~Gool, L.
\newblock Food-101--mining discriminative components with random forests.
\newblock \emph{ECCV}, pp.\  446--461, 2014.

\bibitem[Cai et~al.(2019)Cai, Zhu, and Han]{proxyless18}
Cai, H., Zhu, L., and Han, S.
\newblock Proxylessnas: Direct neural architecture search on target task and
  hardware.
\newblock \emph{ICLR}, 2019.

\bibitem[Chollet(2017)]{xception17}
Chollet, F.
\newblock Xception: Deep learning with depthwise separable convolutions.
\newblock \emph{CVPR}, pp.\  1610--02357, 2017.

\bibitem[Cubuk et~al.(2019)Cubuk, Zoph, Mane, Vasudevan, and Le]{autoaugment18}
Cubuk, E.~D., Zoph, B., Mane, D., Vasudevan, V., and Le, Q.~V.
\newblock Autoaugment: Learning augmentation policies from data.
\newblock \emph{CVPR}, 2019.

\bibitem[Elfwing et~al.(2018)Elfwing, Uchibe, and Doya]{swishsil18}
Elfwing, S., Uchibe, E., and Doya, K.
\newblock Sigmoid-weighted linear units for neural network function
  approximation in reinforcement learning.
\newblock \emph{Neural Networks}, 107:\penalty0 3--11, 2018.

\bibitem[Gholami et~al.(2018)Gholami, Kwon, Wu, Tai, Yue, Jin, Zhao, and
  Keutzer]{squeezeNext18}
Gholami, A., Kwon, K., Wu, B., Tai, Z., Yue, X., Jin, P., Zhao, S., and
  Keutzer, K.
\newblock Squeezenext: Hardware-aware neural network design.
\newblock \emph{ECV Workshop at CVPR'18}, 2018.

\bibitem[Han et~al.(2016)Han, Mao, and Dally]{quantize15}
Han, S., Mao, H., and Dally, W.~J.
\newblock Deep compression: Compressing deep neural networks with pruning,
  trained quantization and huffman coding.
\newblock \emph{ICLR}, 2016.

\bibitem[He et~al.(2016)He, Zhang, Ren, and Sun]{resnet16}
He, K., Zhang, X., Ren, S., and Sun, J.
\newblock Deep residual learning for image recognition.
\newblock \emph{CVPR}, pp.\  770--778, 2016.

\bibitem[He et~al.(2017)He, Gkioxari, Doll{\'a}r, and Girshick]{maskrcnn17}
He, K., Gkioxari, G., Doll{\'a}r, P., and Girshick, R.
\newblock Mask r-cnn.
\newblock \emph{ICCV}, pp.\  2980--2988, 2017.

\bibitem[He et~al.(2018)He, Lin, Liu, Wang, Li, and Han]{amc18}
He, Y., Lin, J., Liu, Z., Wang, H., Li, L.-J., and Han, S.
\newblock Amc: Automl for model compression and acceleration on mobile devices.
\newblock \emph{ECCV}, 2018.

\bibitem[Hendrycks \& Gimpel(2016)Hendrycks and Gimpel]{gelu16}
Hendrycks, D. and Gimpel, K.
\newblock Gaussian error linear units (gelus).
\newblock \emph{arXiv preprint arXiv:1606.08415}, 2016.

\bibitem[Howard et~al.(2017)Howard, Zhu, Chen, Kalenichenko, Wang, Weyand,
  Andreetto, and Adam]{mobilenetv117}
Howard, A.~G., Zhu, M., Chen, B., Kalenichenko, D., Wang, W., Weyand, T.,
  Andreetto, M., and Adam, H.
\newblock Mobilenets: Efficient convolutional neural networks for mobile vision
  applications.
\newblock \emph{arXiv preprint arXiv:1704.04861}, 2017.

\bibitem[Hu et~al.(2018)Hu, Shen, and Sun]{senet18}
Hu, J., Shen, L., and Sun, G.
\newblock Squeeze-and-excitation networks.
\newblock \emph{CVPR}, 2018.

\bibitem[Huang et~al.(2016)Huang, Sun, Liu, Sedra, and Weinberger]{droppath16}
Huang, G., Sun, Y., Liu, Z., Sedra, D., and Weinberger, K.~Q.
\newblock Deep networks with stochastic depth.
\newblock \emph{ECCV}, pp.\  646--661, 2016.

\bibitem[Huang et~al.(2017)Huang, Liu, Van Der~Maaten, and
  Weinberger]{densenet17}
Huang, G., Liu, Z., Van Der~Maaten, L., and Weinberger, K.~Q.
\newblock Densely connected convolutional networks.
\newblock \emph{CVPR}, 2017.

\bibitem[Huang et~al.(2018)Huang, Cheng, Chen, Lee, Ngiam, Le, and
  Chen]{gpipe18}
Huang, Y., Cheng, Y., Chen, D., Lee, H., Ngiam, J., Le, Q.~V., and Chen, Z.
\newblock Gpipe: Efficient training of giant neural networks using pipeline
  parallelism.
\newblock \emph{arXiv preprint arXiv:1808.07233}, 2018.

\bibitem[Iandola et~al.(2016)Iandola, Han, Moskewicz, Ashraf, Dally, and
  Keutzer]{squeezenet16}
Iandola, F.~N., Han, S., Moskewicz, M.~W., Ashraf, K., Dally, W.~J., and
  Keutzer, K.
\newblock Squeezenet: Alexnet-level accuracy with 50x fewer parameters and
  $<$0.5 mb model size.
\newblock \emph{arXiv preprint arXiv:1602.07360}, 2016.

\bibitem[Ioffe \& Szegedy(2015)Ioffe and Szegedy]{batchnorm15}
Ioffe, S. and Szegedy, C.
\newblock Batch normalization: Accelerating deep network training by reducing
  internal covariate shift.
\newblock \emph{ICML}, pp.\  448--456, 2015.

\bibitem[Kornblith et~al.(2019)Kornblith, Shlens, and Le]{imagenettransfer18}
Kornblith, S., Shlens, J., and Le, Q.~V.
\newblock Do better imagenet models transfer better?
\newblock \emph{CVPR}, 2019.

\bibitem[Krause et~al.(2013)Krause, Deng, Stark, and Fei-Fei]{stanfordcars}
Krause, J., Deng, J., Stark, M., and Fei-Fei, L.
\newblock Collecting a large-scale dataset of fine-grained cars.
\newblock \emph{Second Workshop on Fine-Grained Visual Categorizatio}, 2013.

\bibitem[Krizhevsky \& Hinton(2009)Krizhevsky and Hinton]{cifar}
Krizhevsky, A. and Hinton, G.
\newblock Learning multiple layers of features from tiny images.
\newblock \emph{Technical Report}, 2009.

\bibitem[Krizhevsky et~al.(2012)Krizhevsky, Sutskever, and Hinton]{alexnet12}
Krizhevsky, A., Sutskever, I., and Hinton, G.~E.
\newblock Imagenet classification with deep convolutional neural networks.
\newblock In \emph{NIPS}, pp.\  1097--1105, 2012.

\bibitem[Lin \& Jegelka(2018)Lin and Jegelka]{expressoneneuron18}
Lin, H. and Jegelka, S.
\newblock Resnet with one-neuron hidden layers is a universal approximator.
\newblock \emph{NeurIPS}, pp.\  6172--6181, 2018.

\bibitem[Lin et~al.(2017)Lin, Doll{\'a}r, Girshick, He, Hariharan, and
  Belongie]{retinanet17}
Lin, T.-Y., Doll{\'a}r, P., Girshick, R., He, K., Hariharan, B., and Belongie,
  S.
\newblock Feature pyramid networks for object detection.
\newblock \emph{CVPR}, 2017.

\bibitem[Liu et~al.(2018)Liu, Zoph, Shlens, Hua, Li, Fei-Fei, Yuille, Huang,
  and Murphy]{pnas18}
Liu, C., Zoph, B., Shlens, J., Hua, W., Li, L.-J., Fei-Fei, L., Yuille, A.,
  Huang, J., and Murphy, K.
\newblock Progressive neural architecture search.
\newblock \emph{ECCV}, 2018.

\bibitem[Lu et~al.(2018)Lu, Pu, Wang, Hu, and Wang]{expresswidth18}
Lu, Z., Pu, H., Wang, F., Hu, Z., and Wang, L.
\newblock The expressive power of neural networks: A view from the width.
\newblock \emph{NeurIPS}, 2018.

\bibitem[Ma et~al.(2018)Ma, Zhang, Zheng, and Sun]{shufflenetv218}
Ma, N., Zhang, X., Zheng, H.-T., and Sun, J.
\newblock Shufflenet v2: Practical guidelines for efficient cnn architecture
  design.
\newblock \emph{ECCV}, 2018.

\bibitem[Mahajan et~al.(2018)Mahajan, Girshick, Ramanathan, He, Paluri, Li,
  Bharambe, and van~der Maaten]{imagenetinstagram18}
Mahajan, D., Girshick, R., Ramanathan, V., He, K., Paluri, M., Li, Y.,
  Bharambe, A., and van~der Maaten, L.
\newblock Exploring the limits of weakly supervised pretraining.
\newblock \emph{arXiv preprint arXiv:1805.00932}, 2018.

\bibitem[Maji et~al.(2013)Maji, Rahtu, Kannala, Blaschko, and
  Vedaldi]{aircraft}
Maji, S., Rahtu, E., Kannala, J., Blaschko, M., and Vedaldi, A.
\newblock Fine-grained visual classification of aircraft.
\newblock \emph{arXiv preprint arXiv:1306.5151}, 2013.

\bibitem[Ngiam et~al.(2018)Ngiam, Peng, Vasudevan, Kornblith, Le, and
  Pang]{domainjft18}
Ngiam, J., Peng, D., Vasudevan, V., Kornblith, S., Le, Q.~V., and Pang, R.
\newblock Domain adaptive transfer learning with specialist models.
\newblock \emph{arXiv preprint arXiv:1811.07056}, 2018.

\bibitem[Nilsback \& Zisserman(2008)Nilsback and Zisserman]{flowers}
Nilsback, M.-E. and Zisserman, A.
\newblock Automated flower classification over a large number of classes.
\newblock \emph{ICVGIP}, pp.\  722--729, 2008.

\bibitem[Parkhi et~al.(2012)Parkhi, Vedaldi, Zisserman, and
  Jawahar]{oxfordpets}
Parkhi, O.~M., Vedaldi, A., Zisserman, A., and Jawahar, C.
\newblock Cats and dogs.
\newblock \emph{CVPR}, pp.\  3498--3505, 2012.

\bibitem[Raghu et~al.(2017)Raghu, Poole, Kleinberg, Ganguli, and
  Sohl-Dickstein]{expressdepth17}
Raghu, M., Poole, B., Kleinberg, J., Ganguli, S., and Sohl-Dickstein, J.
\newblock On the expressive power of deep neural networks.
\newblock \emph{ICML}, 2017.

\bibitem[Ramachandran et~al.(2018)Ramachandran, Zoph, and Le]{swish18}
Ramachandran, P., Zoph, B., and Le, Q.~V.
\newblock Searching for activation functions.
\newblock \emph{arXiv preprint arXiv:1710.05941}, 2018.

\bibitem[Real et~al.(2019)Real, Aggarwal, Huang, and Le]{amoebanets18}
Real, E., Aggarwal, A., Huang, Y., and Le, Q.~V.
\newblock Regularized evolution for image classifier architecture search.
\newblock \emph{AAAI}, 2019.

\bibitem[Russakovsky et~al.(2015)Russakovsky, Deng, Su, Krause, Satheesh, Ma,
  Huang, Karpathy, Khosla, Bernstein, et~al.]{imagenet15}
Russakovsky, O., Deng, J., Su, H., Krause, J., Satheesh, S., Ma, S., Huang, Z.,
  Karpathy, A., Khosla, A., Bernstein, M., et~al.
\newblock Imagenet large scale visual recognition challenge.
\newblock \emph{International Journal of Computer Vision}, 115\penalty0
  (3):\penalty0 211--252, 2015.

\bibitem[Sandler et~al.(2018)Sandler, Howard, Zhu, Zhmoginov, and
  Chen]{mobilenetv218}
Sandler, M., Howard, A., Zhu, M., Zhmoginov, A., and Chen, L.-C.
\newblock Mobilenetv2: Inverted residuals and linear bottlenecks.
\newblock \emph{CVPR}, 2018.

\bibitem[Sharir \& Shashua(2018)Sharir and Shashua]{expressoverlap18}
Sharir, O. and Shashua, A.
\newblock On the expressive power of overlapping architectures of deep
  learning.
\newblock \emph{ICLR}, 2018.

\bibitem[Srivastava et~al.(2014)Srivastava, Hinton, Krizhevsky, Sutskever, and
  Salakhutdinov]{dropout14}
Srivastava, N., Hinton, G., Krizhevsky, A., Sutskever, I., and Salakhutdinov,
  R.
\newblock Dropout: a simple way to prevent neural networks from overfitting.
\newblock \emph{The Journal of Machine Learning Research}, 15\penalty0
  (1):\penalty0 1929--1958, 2014.

\bibitem[Szegedy et~al.(2015)Szegedy, Liu, Jia, Sermanet, Reed, Anguelov,
  Erhan, Vanhoucke, and Rabinovich]{googlenet14}
Szegedy, C., Liu, W., Jia, Y., Sermanet, P., Reed, S., Anguelov, D., Erhan, D.,
  Vanhoucke, V., and Rabinovich, A.
\newblock Going deeper with convolutions.
\newblock \emph{CVPR}, pp.\  1--9, 2015.

\bibitem[Szegedy et~al.(2016)Szegedy, Vanhoucke, Ioffe, Shlens, and
  Wojna]{inceptionv316}
Szegedy, C., Vanhoucke, V., Ioffe, S., Shlens, J., and Wojna, Z.
\newblock Rethinking the inception architecture for computer vision.
\newblock \emph{CVPR}, pp.\  2818--2826, 2016.

\bibitem[Szegedy et~al.(2017)Szegedy, Ioffe, Vanhoucke, and
  Alemi]{inceptionv417}
Szegedy, C., Ioffe, S., Vanhoucke, V., and Alemi, A.~A.
\newblock Inception-v4, inception-resnet and the impact of residual connections
  on learning.
\newblock \emph{AAAI}, 4:\penalty0 12, 2017.

\bibitem[Tan et~al.(2019)Tan, Chen, Pang, Vasudevan, Sandler, Howard, and
  Le]{mnas18}
Tan, M., Chen, B., Pang, R., Vasudevan, V., Sandler, M., Howard, A., and Le,
  Q.~V.
\newblock {MnasNet}: Platform-aware neural architecture search for mobile.
\newblock \emph{CVPR}, 2019.

\bibitem[Xie et~al.(2017)Xie, Girshick, Doll{\'a}r, Tu, and He]{resnext17}
Xie, S., Girshick, R., Doll{\'a}r, P., Tu, Z., and He, K.
\newblock Aggregated residual transformations for deep neural networks.
\newblock \emph{CVPR}, pp.\  5987--5995, 2017.

\bibitem[Yang et~al.(2018)Yang, Howard, Chen, Zhang, Go, Sze, and
  Adam]{netadapt18}
Yang, T.-J., Howard, A., Chen, B., Zhang, X., Go, A., Sze, V., and Adam, H.
\newblock Netadapt: Platform-aware neural network adaptation for mobile
  applications.
\newblock \emph{ECCV}, 2018.

\bibitem[Zagoruyko \& Komodakis(2016)Zagoruyko and Komodakis]{wideresnet16}
Zagoruyko, S. and Komodakis, N.
\newblock Wide residual networks.
\newblock \emph{BMVC}, 2016.

\bibitem[Zhang et~al.(2017)Zhang, Li, Loy, and Lin]{polynet17}
Zhang, X., Li, Z., Loy, C.~C., and Lin, D.
\newblock Polynet: A pursuit of structural diversity in very deep networks.
\newblock \emph{CVPR}, pp.\  3900--3908, 2017.

\bibitem[Zhang et~al.(2018)Zhang, Zhou, Lin, and Sun]{shufflenet17}
Zhang, X., Zhou, X., Lin, M., and Sun, J.
\newblock Shufflenet: An extremely efficient convolutional neural network for
  mobile devices.
\newblock \emph{CVPR}, 2018.

\bibitem[Zhou et~al.(2016)Zhou, Khosla, Lapedriza, Oliva, and Torralba]{cam16}
Zhou, B., Khosla, A., Lapedriza, A., Oliva, A., and Torralba, A.
\newblock Learning deep features for discriminative localization.
\newblock \emph{CVPR}, pp.\  2921--2929, 2016.

\bibitem[Zoph \& Le(2017)Zoph and Le]{nas_cifar17}
Zoph, B. and Le, Q.~V.
\newblock Neural architecture search with reinforcement learning.
\newblock \emph{ICLR}, 2017.

\bibitem[Zoph et~al.(2018)Zoph, Vasudevan, Shlens, and Le]{nas_imagenet18}
Zoph, B., Vasudevan, V., Shlens, J., and Le, Q.~V.
\newblock Learning transferable architectures for scalable image recognition.
\newblock \emph{CVPR}, 2018.

\end{thebibliography}

\end{document}